%% file: main.tex
\documentclass[twocolumn]{svjour3}          
\usepackage{times}
\usepackage[pdftex,colorlinks,citecolor=blue, urlcolor=blue,]{hyperref}
\usepackage[numbered]{bookmark}
\usepackage{tabularx}
\usepackage{amsmath,amssymb} 
\usepackage{multirow}
\usepackage{listings}
\usepackage{graphicx}
\usepackage{subfigure}
\usepackage{caption}
\usepackage{booktabs} 
\usepackage{xcolor,colortbl}
\usepackage{bm}
\usepackage{natbib}

\begin{document}
\title{Zoom Out-and-In Network 
	with Map Attention Decision \\for Region Proposal and Object Detection}

\author{Hongyang Li        \and
        Yu Liu      \and
        Wanli Ouyang       \and
        Xiaogang Wang 
}


\institute{H. Li \and Y. Liu \and X. Wang \at
	Department of Electronic Engineering\\
	The Chinese University of Hong Kong, Hong Kong, China \\
	\email{\texttt{\{yangli,yuliu,xgwang\}@ee.cuhk.edu.hk} } \and
	W. Ouyang \at 
	University of Sydney, Sydney, Australia\\
	\email{\texttt{wanli.ouyang@sydney.edu.au}}}

\date{Received: date / Accepted: date}

\maketitle

\begin{abstract}
In this paper, we propose a zoom-out-and-in network for generating object proposals. 
A key observation is that 
it is difficult to classify anchors of different sizes with the same set of features.
Anchors of different sizes should be placed accordingly based on different depth within a network: smaller boxes on high-resolution layers with a smaller stride 
while larger boxes on low-resolution counterparts with a larger stride.
Inspired by the conv/deconv structure, we fully leverage the low-level local details and high-level regional semantics from two feature map streams, which are complimentary to each other, to identify the objectness in an image.
A map attention decision (MAD) unit is further proposed to aggressively search for neuron activations among two streams and attend the most 
contributive ones on the feature learning of the final loss. The unit serves as a decision-maker to adaptively activate maps along certain channels 
with the solely purpose of optimizing the overall training loss.
%
One advantage of MAD is that 
%
the learned weights enforced on each feature channel is predicted on-the-fly based on the input context, 
which is more suitable than the fixed enforcement of a convolutional kernel.
Experimental results on three datasets, including PASCAL VOC 2007, ImageNet DET, MS COCO, demonstrate the effectiveness of our proposed algorithm over other state-of-the-arts, in terms of average recall (AR) for region proposal and average precision (AP) for object detection.
\end{abstract}

%
%

\input{intro_v1.tex}
\input{related.tex}

\input{alg.tex}

\input{experiments.tex}
\section{Conclusions}\label{sec:conclusions}

In this work, we devise a zoom-out-and-in network 
that both utilizes  low-level details and high-level semantics. 
The
information from top layers is gradually up-sampled by deconvolution to reach suitable resolution for small-sized objects.  
Such a strategy could alleviate the drawback of identifying small objects on feature maps with a large stride.
We further propose a map attention decision (MAD) unit to actively search for neuron activations and attend to specific maps that could weigh more during the feature learning in the final RPN layer. Several training strategies are also proposed and investigated to enhance the quality of region proposals. 
Experiments for both the region proposal generation and object detection tasks show that our proposed algorithm (ZIP with map attention decision) performs superior against previous state-of-the-arts on  popular benchmarks, including PASCAL VOC 2007, ImageNet DET and MS COCO.
%

\section*{Acknowledgment}
We would like to thank   reviewers for helpful comments, 
S. Gidaris, X. Tong and K. Kang
for fruitful discussions along the way, W. Yang for proofreading
the manuscript. H. Li is funded by the Hong Kong
Ph.D. Fellowship scheme. We are also grateful for SenseTime
Group Ltd. donating the resource of GPUs at time of this project.

\bibliographystyle{apalike}      
\bibliography{deep_learning,obj_prop}   
\end{document}

%% file: intro_v1.tex
\section{Introduction}
\label{Sec:intro}

Object proposal is the task of proposing a set of candidate regions or bounding boxes in an image that may potentially contain an object.
In recent years, the emergence of object proposal algorithms \citep{selective_search,prime,MCG,co_generate,hyper_net,deep_proposal,object_proposal_eval,pronet,li2017zoom}
have significantly boosted the development of many vision tasks,
\citep{liu2017learning,liu2017recurrent,li2016multi,czz_tracking,li2017we}, 
especially for  object detection \citep{rcnn,rfcn,fast_rcnn,inside_outside,ssd}.
It is verified by Hosang \textit{et.al} \citep{Hosang2015Pami} that region proposals with high average recall correlates well with good performance of a detector. Thus generating object proposals has quickly become the de-facto pre-processing step.

Currently, CNN models are known to be effective in generating candidate boxes \citep{faster_rcnn,deep_box,hyper_net}. Existing works use deep CNN features at the last layer for classifying whether a candidate box should be an object proposal. The candidate box can come from random seed \citep{attractioNet}, external boxes (selective search \citep{selective_search}, edge box \citep{edgebox}, \textit{etc.}), or 
sliding windows \citep{overfeat}. Deep CNN-based proposal methods employ a zoom-out network, where down-sampling is used for reducing the resolution of features. 
This zoom-out design is good for image classification since 
down-sampling is effective for achieving translation invariance, increasing the receptive field of features, and saving computation.

\begin{figure}
	\begin{center}
		\includegraphics[width=0.42\textwidth]{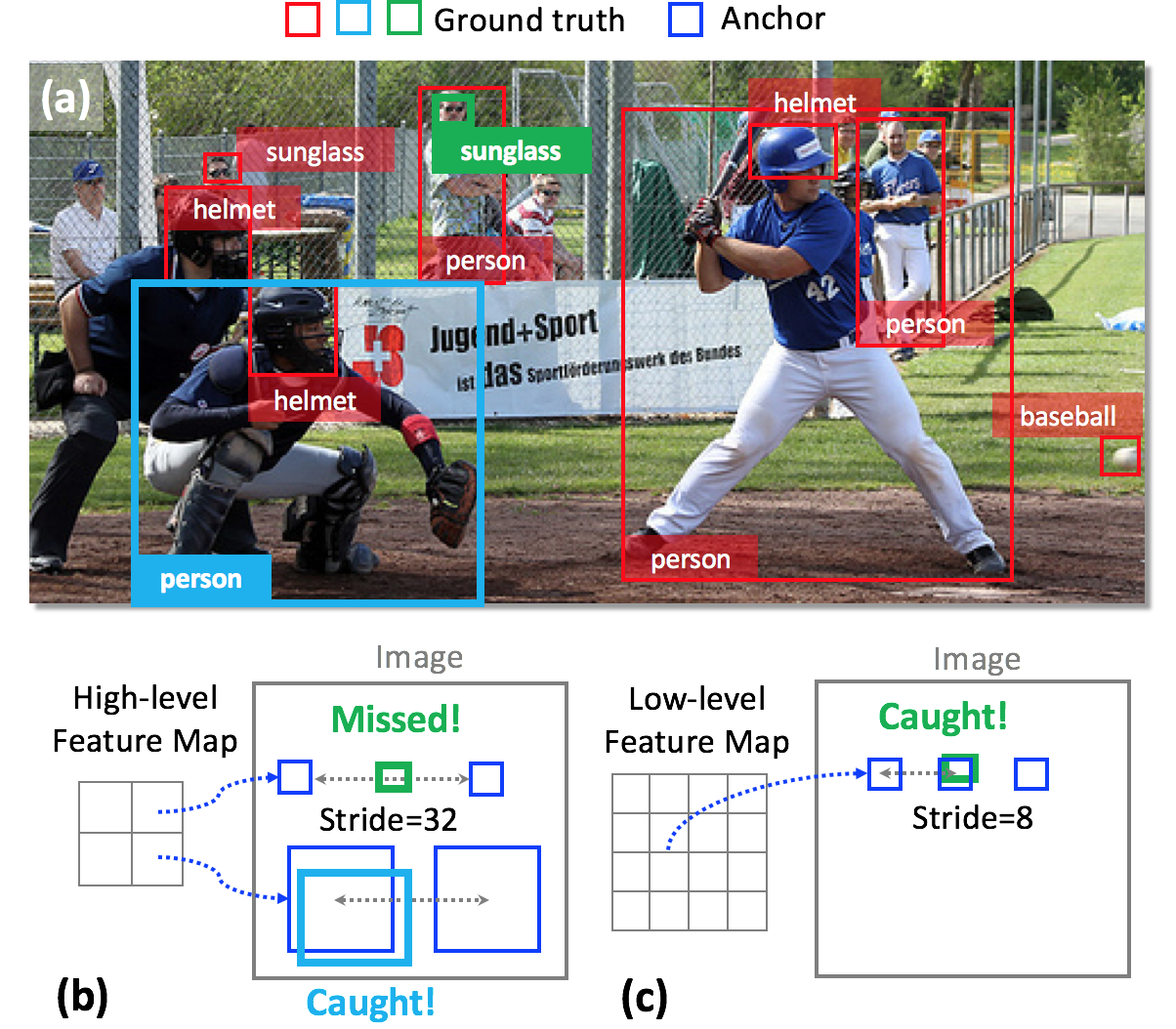}
	\end{center}
	\vspace{-.4cm}
	\caption{
		(a) Input image where blue and green objects are being examined. (b) A larger stride in higher layers miss the green box between two adjacent anchors. (c) This motivates us to place small-scale anchors in the preceding layers with a smaller stride.
	}
	\label{fig:motivation}
\end{figure}

However, we argue that the zoom out structure faces great challenge by using the same set of features and the same classifier to handle object proposals in different sizes.
The learned features have to sacrifice on large objects in order to compromise small ones. It also has two problems for detecting small objects.
%
%
First, candidate anchors are placed at the final feature map in existing works \citep{faster_rcnn}. 
As shown in Figure~\ref{fig:motivation}(b),
when the down-sampling rate (or total stride) of the feature map is 32, moving the anchor by one step on the feature map corresponds to stepping by 32 pixels in the image.  
Anchors might skip small objects due to a larger stride. 
Second, for the case of small objects, the down-sampling operation makes the network difficult to determine 
whether neurons or features will be activated or not for the subsequent layer.
The lack of resolution in feature map is a factor that influences the ability of object proposal methods in finding small objects. 
As depicted in Figure~\ref{fig:motivation}(c), if the resolution of feature map is sufficient, the anchor box with smaller stride can locate small objects. 
%
%
%
To fit for the zoom-out network design, one could leverage features at shallow layers which has higher resolution. 
Features from shallow layers are yet weak in extracting high-level information that is essential for object proposal. 
%
It would be desirable if  high-level semantics can pass information to guide  feature learning in the lower counterparts for small objects.

\begin{figure}[t]
	\begin{center}
		\includegraphics[width=.48\textwidth]{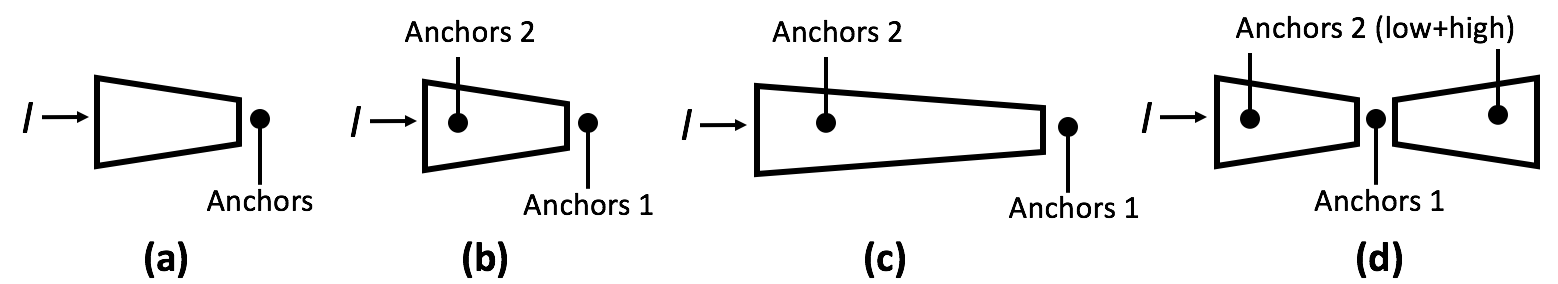}
	\end{center}
	\vspace{-.4cm}
	\caption{Different network design for region proposals.
		(a) RPN \citep{faster_rcnn}; (b) RPN with split anchors; 
		(c) Deeper RPN with split anchors; 
		(d) Zoom out-and-in network adopted in this paper.
	}
	\label{fig:structure}
\end{figure}

Inspired by the conv/deconv structure \citep{hg,long2014fully}, we propose a \textit{\textbf{Z}}oom-out-and-\textit{\textbf{I}}n network for object \textit{\textbf{P}}roposals (ZIP) to both utilize feature maps from high and low streams.
Figure \ref{fig:structure} shows the network structure of different designs at a glance.
Under such a scheme not only the detection of small objects enhances, but also the detection of larger anchors increases.
%
%
By splitting anchors of different size, we make each level have its own classifier  and own set of features to handle a specific range of scales. 

\begin{figure*}
	\begin{center}
		\includegraphics[width=.95\linewidth]{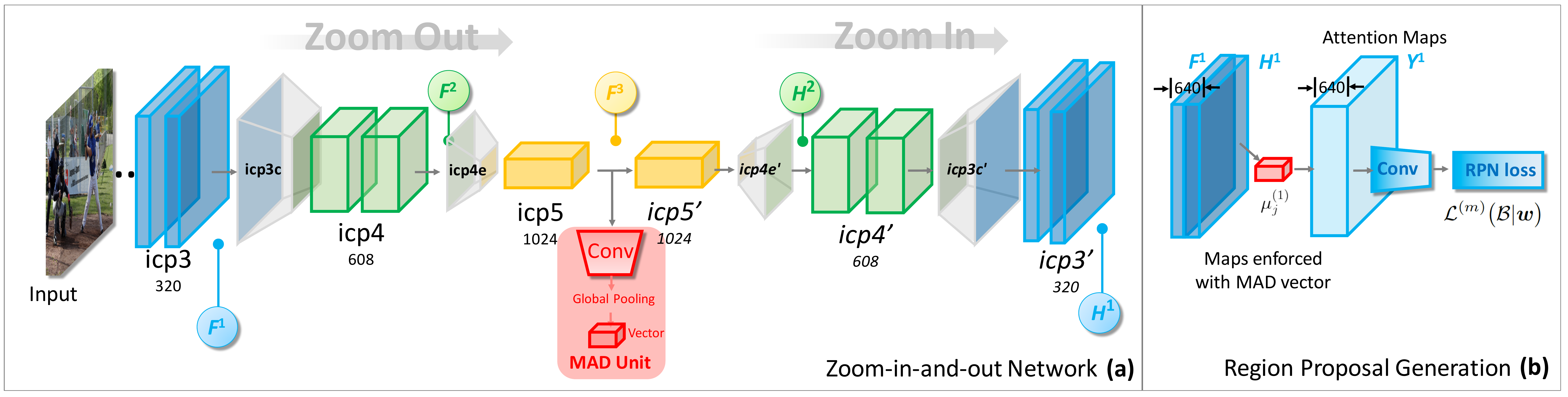}
	\end{center}
	\vspace{-.3cm}
	\caption{(a) Zoom network with map attention decision (MAD) unit. 
		Prior to down-sampling in each block, we utilize the feature maps at that depth to place different size of anchors and identify the objectness in the image.
		Feature maps 
		are actively determined by the MAD unit,
		using a selective vector $\mu$ to decide the attention of neurons from both low-level and high-level layers.
		(b) The region proposal generation pipeline after the zoom network. It only shows the case on level $m=1$, where maps $\bm{F}^1, \bm{H}^1$ from two streams, are leveraged by a MAD vector to generate the attention maps $\bm{Y}^1$. Level $m=2$ shares similar workflow as $m=1$ whilst level $m=3$ neither has merging from different sources nor applies MAD.
			The complete network diagram for region proposal and object detection is depicted in Figure \ref{fig:pipeline_det}.
	}
	\label{fig:pipeline}
\end{figure*}

In this paper, we devise a \textbf{\textit{M}}ap \textbf{\textit{A}}ttention \textbf{\textit{D}}ecision unit (MAD), to actively search for neuron activations among feature maps from low-level and high-level streams.
Figure \ref{fig:pipeline} illustrates the pipeline of our algorithm.
MAD unit is a side branch taking the feature maps in the highest level as input and generates a column vector, whose length corresponds to two times as the number of channels in maps that MAD enforces on.
The weight vector evaluates the contribution of each map  on the feature learning in the final RPN loss layer. Such an intuition is derived from the gradient of loss with respect to MAD unit (see Section \ref{sec:training-with-recursive-regression}).
The proposed unit involves parameter update during training
and yet is relatively isolated from the weights of the network.
In this sense, it resembles an external memory in neural networks \citep{Graves_Nature2016,center_loss}.
One advantage of MAD in selecting neuron activations among feature maps is that it considers the context of input - the weight vector is computed on the fly for each training or test sample, which shares a big distinction in previous work where weights in the learned combination kernel are fixed (see Section \ref{sec:related-work}).

Our algorithm is implemented in  Caffe \citep{caffe}  with Matlab wrapper and trained on multiple GPUs. 
The codebase and object proposal results of our method 
are available online\footnote{\href{https://github.com/hli2020/zoom_network}{\texttt{https://github.com/hli2020/zoom\_network}}}.
To sum up, our contributions in this work are as follows:
\begin{enumerate}
	\item A map attention decision unit to actively search for neuron activations among feature maps and attend the most contributive ones on the feature learning of the loss. It adapts the weights among feature maps on the fly based on the context of input.
	
	\item A zoom-out-and-in network that utilizes 
	feature maps at different depth of a network to leverage both low-level details and high-level semantics.
	Anchors are placed separately based on different resolutions of maps to detect objects of various sizes.
	
	\item Several practical techniques for generating effective proposals and enhancing the object detection performance, including recursive test and training, boosting the classifiers, etc., are investigated. 
	The proposed ZIP algorithm achieves average recall (AR) to 68.8\% and 61.2\% at top 500 proposals on ILSVRC DET and MS COCO, respectively. 
	Furthermore, the proposed boxes will improve average precision (AP) by around 2\% for object detection compared to previous state-of-the-art. 
	
\end{enumerate}

The rest of the paper is organized as follows: Section \ref{sec:related-work} states the related work from four aspects: the spirit of utilizing various information from different levels (depth or  resolution) in the network, conv/deconv structure walkthrough, region proposal and object detection. Section \ref{sec:sop-algorithm} and \ref{sec:zoom-network-with-recursive-training-for-object-detection} depict the detailed description on each component of the proposed framework for region proposal and object detection, respectively. Section \ref{experiments} verifies the effectiveness of ZIP algorithm by experiments. Section \ref{sec:conclusions} concludes the paper.

%% file: related.tex
\section{Related Work}\label{sec:related-work}

\textbf{Using features from multi-depth in the network.}
The idea of utilizing different feature maps at different locations in the network has been investigated and proved to be effective \citep{ssd,SharpMask,pyramid_feature_learn}. 
%
%
Pinheiro \textit{et al.} \citep{SharpMask} proposed a network to fully leverage all feature maps. Instead of simply combing all feature maps, they devised a upsampling path to gradually refined maps that contained high-level semantics.
SSD \citep{ssd} 
considers the idea of putting different scales of anchors on various depth within the network.
%
However, SSD inherently does not consider the zoom-in (deconvolution) structure to help guide the learning in lower layers. 
Lin \textit{et al.} \citep{pyramid_feature_learn} introduced a feature pyramid network which up-samples high-level features and generates multiple predictions.
However, the top-down path is achieved by convolution \textit{kernels}: once the training is finished, weights are fixed for test; in ours, the decision of neuron activations are determined by the input's context, \textit{i.e.}, based on the predictions in the MAD vector.

\textbf{Conv/deconv structure.} The spirit of upsampling feature maps through learnable parameters is known as deconvolution, which is widely applied in other vision domains \citep{long2014fully,noh_iccv15, hypercolumn, hg}. 
Shelhamer \textit{et. al} \citep{long2014fully} first devise a novel structure to do pixel-wise semantic segmentation via learnable deconvolution.
In \citep{u_net}, a U-shaped network for segmentation is designed with a contracting path to capture context and a symmetric expanding path to localize objects.
Newell \textit{et. al} \citep{hg} proposed an 
hourglass structure 
where feature maps are passed with skip connections through stacks for pose estimation.
%
DSSD \citep{dssd}  embedded the original  SSD framework into 
a similar U-net structure with more discriminative
features
due to the up-sampling subnetwork;
and more sophisticated prediction modules are designed for each detection branch.
Our work is inspired by these works and yet have clear distinctions besides different application domain.
Existing approaches  concatenate all features \citep{hypercolumn}
 or use the final feature map for prediction  \citep{hg}, 
 while we utilize specific features at different locations of a network to detect objects of various size.
With such a philosophy in mind, we have each object equipped with suitable features at a proper resolution.

\textbf{Region proposals.} In early stages people resort to finding the objectness via multiple cues and larger pixels (called superpixel) in a semantic manner \citep{selective_search,objectness,gop}.
In \citep{learn_prop}, a learning method is proposed by training an ensemble
of figure-ground segmentation models jointly, where 
individual models can specialize and complement each other.
In recent years, CNN-based approaches \citep{co_generate,deep_proposal,from_p2c,orient_prop} are more popular with a non-trivial margin of performance boost.
Jie \textit{et al.} \citep{scale_aware} proposed a scale-aware pixel-wise proposal framework where two separate networks are learned to handle large and small objects, respectively.
Pinheiro \textit{et al.} \citep{deepMask} devised a discriminative CNN model to generate a mask and predict its likelihood of containing an object.
In \citep{attractioNet}, a refine-and-repeat model is formulated to recursively refine the box locations based the revised score from a trained model.

\textbf{Object detection.} Our main pipeline follows the popular object detection prototype \citep{rcnn,fast_rcnn,faster_rcnn,ssd,yolo_v1,yolo_v2} of using region proposals as initial point to localize objects and classify them.
It is found out that the best practice for detection is to have a set of object boxes passing a RoI-pooling layer, where afterwards we feed the fix-sized feature maps into several additional convolutional layers \citep{resNet,spp}. The final loss is similar to that in region proposal: a classification score and a box location adjustment via regression. Recently, the community has witnessed more advanced versions of the Faster-RCNN detector and its variants. \citep{rfcn} took a further step to move the functionality of RoI-pooling to the very last feature layer before the classification: the scheme of position-sensitive score maps addresses the dilemma between 
invariance in classification and variance in detection under a region-based, fully convolutional network. 
Dai \textit{et al.} \citep{deformable} motivated that the receptive field of convolutional kernels should also be conditioned on the input RoI boxes and thus formulated the offset reception in an end-to-end learning manner; the deformable unit could be plugged into most popular network structures. 
Wang \textit{et al.} \citep{a_fast_rcnn} proposed to learn an adversarial network that generates examples with occlusions and deformations, which are hard for the detector to classify. These examples are rare in long-tail distribution and the adversarial part could model such an occurrence.

%% file: alg.tex
\section{ZIP with Map Attention for Region Proposal}\label{sec:sop-algorithm}
\subsection{Backbone Architecture}\label{sec:network-architecture}

Figure \ref{fig:pipeline} describes the overview of the proposed zoom out-and-in network with MAD.
Most existing network structures \citep{alexnet,resNet,vgg,faster_rcnn,li2016multi} can be viewed and used as a zoom-out network. 
We adopt the Inception-BN  model \citep{bn} and use the inception module as basic block throughout the paper. Specifically, 
an image is first fed into three convolutional layers, after which the feature maps are downsampled by a total stride of 8. There are nine inception modules afterwards, denoted as 
\texttt{icp3a-3c}, \texttt{icp4a-4e} and \texttt{icp5a-5b}. Max-pooling is placed after \texttt{icp3c} and \texttt{icp4e}. 
Therefore, we divide the network into three parts based on the stride at 8, 16 and 32. Denote $\bm{F}^{(m)}$ at depth (or level) $m$ as the output feature maps, where $m=1,2,3$ is the depth index. Note that the spatial size in $\bm{F}^{(m)}$ is two times the size of that in its subsequent level, which is determined by our network design.
The number of feature channels in each  block
(from \texttt{icp3} to \texttt{icp5}) are 320, 608 and 1024, respectively. 

Inspired by the conv-deconv network in other vision domains \citep{hg,noh_iccv15}, we adopt a zoom-in  architecture to better 
leverage the summarized high-level feature maps for refining its low-level counterparts.
Such a zoom-in architecture is exactly the mirrored version of the zoom-out part with max-pooling being replaced by deconvolution.
We denote the mirrored inception block
as \texttt{icpX'}. It is found in preliminary experiments that a bilinear upsampling plus a convolution operation achieves better performance than that of an existing deconvolution layer in practice. Denote the output feature maps after the up-sampling inceptional layer \texttt{icp4e'} as
$\bm{H}^{(2)}$ and the output maps at layer \texttt{icp3'} as $\bm{H}^{(1)}$. The neuron activations in these feature maps contain the high-level regional semantics, summarized via the zoom-out network and propagated back to guide the feature learning in low-level layer via deconvolution layer by layer.
Note that we do not deliberately formulate a $\bm{H}^{(3)}$   since $\bm{F}^{(3)}$ alone are suitable enough for identifying large objects.

\subsection{Zoom Network Training with MAD Unit}\label{sec:training-with-recursive-regression}

\begin{figure*}
	\begin{center}
		\includegraphics[width=.85\textwidth]{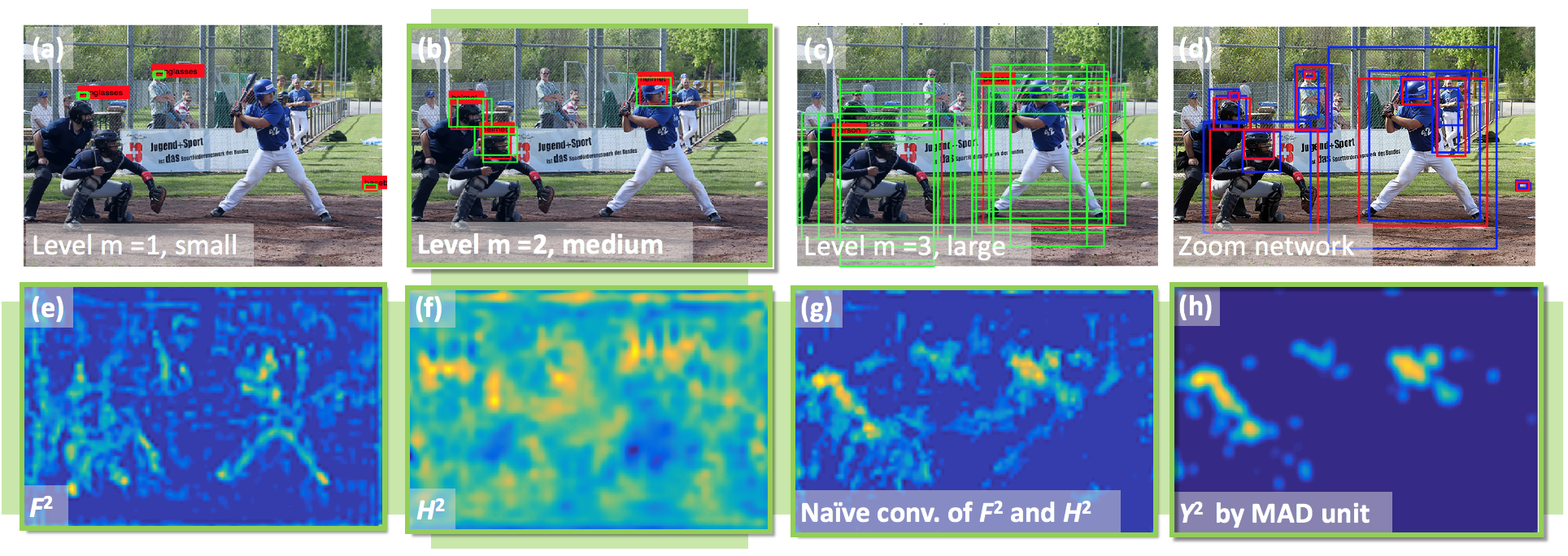}
	\end{center}
	\vspace{-.4cm}
	\caption{Effect of MAD unit in learning crucial attention information of features. 
		(a)-(c): Anchor templates (green boxes) with different scales and aspect ratios are placed at different levels in the network, where ground truth annotations are marked in red. 
		(d): Our proposal results (in blue).
		(e)-(h): Lower neuron activation in $\bm{F}^2$; Higher neuron activation in $\bm{H}^2$; naive convolution output from $\bm{F}^2$ and $\bm{H}^2$; attention map using MAD unit.
		%
	}
	\label{fig:featMap}
\end{figure*}

At the core of the zoom out-and-in network is a feature map attention unit to deterministically choose neuron activations among channels from both high and low level streams, and to help supervise the feature learning in the final RPN layer to identify objectness in the image.
To this end, we propose a feature map selection unit, called MAD, to actively help find useful neuron activations among feature channels (see red area in Figure \ref{fig:pipeline}).

The unit $\texttt{MAD}(\cdot)$ is fed with the feature maps $\bm{F}^3$ as input, goes through three convolutions with kernel size of 3, stride of 1 and the number of channels being exactly two times as that on level $m$ at each layer, where $m=1,2$.
Since the zoom network is a fully convolution design and the size of input image varies, the spatial size of output after the unit also changes. Therefore, we take a  global max-pooling operation 
to dynamically alters the spatial size to $1\times 1$.
We denote the output of the unit on level $m$ as MAD vector:
\begin{equation}
\bm{\mu}^{(m)} = \texttt{MAD}(\bm{F}^3, m) \in \mathbb{R}^{ 1 \times 1 \times 2C^{(m)}}, \label{MAD_basic}
\end{equation}
where 
$C^{(m)}$ is the channel number 
of the input  maps.

Given the inputs from two streams, we first concatenate and denote them as $\bm{X}^{(m)} = [
\bm{F}^{(m)}, \bm{H}^{(m)}  ]$. Then we have the attention maps $\bm{Y}^{(m)}$, which is an outcome of the concatenated maps multiplied by the MAD vector:
\begin{gather}
\bm{y}_j^{(m)} = \mu_j^{(m)} \bm{x}_j^{(m)},
\end{gather}
where 
%
$ \bm{x}_j,  \bm{y}_j$ are the vector representation of $\bm{X}^{(m)}, \bm{Y}^{(m)}$  in channel $j$, respectively.
%
Note that $\mu_j^{(m)}$ is a learning variable and also involves the parameter update rule. 
	There are many possible alternative designs of MAD unit besides the one stated above and we have included an ablative investigation in the experiments (see Section \ref{sec:ablation-study}).

Now the key question is how to derive the gradient of MAD vector and how to interpret its functionality. For simplicity, we drop 
notation $m$ and use the vector form of $\bm{X}$ and $\bm{Y}$ in the following discussion.
Suppose we have the upper gradient $ \nabla_{\bm{y}_j} \mathcal{L}$ flowing back from the loss 
layer, the gradients with respect to the MAD vector and input 
are achieved by chain rule\footnote{The first row is the inner product of two vectors, resulting in a scalar gradient; while the second is the common vector multiplication by a scalar, resulting in a vector also.}:
\begin{align}
\frac{\partial \mathcal{L}}{  \partial \mu_j} = & ~ \nabla_{\bm{y}_j} \mathcal{L} \cdot \frac{\partial \bm{y}_j}{\partial \mu_j} = \nabla_{\bm{y}_j} \mathcal{L}\cdot \bm{x}_j, \label{MAD_grad}\\
\frac{\partial \mathcal{L}}{ \partial \bm{x}_j} =  & ~ \nabla_{\bm{y}_j} \mathcal{L} \cdot \frac{\partial \bm{y}_j}{\partial \bm{x}_j}  = \nabla_{\bm{y}_j} \mathcal{L} \cdot \mu_j.  \label{MAD_2}
\end{align}
The update rule of $\mu_j$ and $\bm{x}_j$ are defined as follows:
\begin{equation}
\mu_j = \mu_j - \alpha \frac{\partial \mathcal{L}}{\partial \mu_j}, ~~\bm{x}_j = \bm{x}_j - \alpha \frac{\partial \mathcal{L}}{\partial \bm{x}_j},
\end{equation}
where $\alpha > 0$ is the learning rate in the unit.

\textbf{Analysis.} 
The gradient $\nabla_{\bm{y}_j} \mathcal{L}$ is the propagated error 
from the final loss. If the data and parameters of the network follow the update rule, the loss decreases and we say the features are well learned.
Keep in mind that weights or data in the network can increase or decrease as long as such an alternation optimizes the loss, or equivalently, benefits the feature learning.
%
If the direction of map $\bm{x}_j$ 
\textit{matches} the upper gradient\footnote{Direction `matches' means the included angle between two vectors in multi-dimensional space is within 90 degrees; and `departs' means the angle falls at $[90, 180]$.}, 
$\nabla_{\mu_j} \mathcal{L}$ is positive with $\mu_j$ {decreased}, and based on the update rule of $\bm{x}_j$, it remains a relatively {high} value, meaning the features does not alter much during update; 
similarly if the direction of $\bm{x}_j$ \textit{departs} from the upper gradient, 
$\nabla_{\mu_j} \mathcal{L}$ is negative with $\mu_j$ {increased}, and $\bm{x}_j$ changes to a relatively {low} value since $\nabla_{\bm{x}_j} \mathcal{L}$ is large.
Therefore, MAD vector $\mu_j$ serves as an {{adapter}} to alter the magnitude within the network: feature maps whose direction are in accordance with the gradient (thus the optimization goal) should remain unchanged; feature maps whose direction are different from the gradient should alter quickly and remain at a low state (inactive neuron).
%
%
%
One advantage of the MAD unit is that it merges feature maps from two streams based on the context of input image, \textit{i.e.}, the vector $\mu_j$ are on-the-fly predictions of the network, whereas in previous work, \textit{e.g.}, \citep{pyramid_feature_learn}, they opt to fixed kernel parameters in the convolutional layer to combine features from different channels during test (see Section \ref{sec:related-work}). 

	The  squeeze-excitation network (SENet) \citep{SENet}
	demises a block module that first squeezes the spatial size of feature maps along the channel dimension, then extracts the summarized information (adaptive recalibration) and at last imposes the output vector of excitation on the feature maps. The recalibration output resembles the MAD unit.
	However, the biggest difference lies in the source where MAD, called ``excitation recalibration vector'' in \citep{SENet}, comes from. In their scheme, the map attention decision vector is generated from the very feature map itself; whereas in our design, the MAD decision comes from the higher feature maps (level $m=3$) to guide or supervise the feature importance learning in the lower layers (on level $m=1$ or $2$). We deem SENet and our proposed one as concurrent works.

\textbf{Training loss.} After obtaining the attention maps from MAD, we further feed them into a subsequent convolution layer and conduct the anchor detection in the final RPN layer.
%
Let $\mathcal{L}^{(m)}\big(  \mathcal{B}   \{  \bm{y}_j  (i)  \} | \bm{w} \big)$ be the RPN loss on depth $m$, 
where $\mathcal{B}$ is the mini-batch;
$\bm{w}$ denotes the weights in the final RPN layer. 
$\bm{y}_j  (i) = \{  y_q (i)  \in \mathbb{R}^{jd} \} $ denotes the $i$-th sample from preceding layer. We extend the vector form of 
$\bm{y}_j$ to the element-wise notation $y_q$ for gradient derivation later, \textit{i.e.}, $q=jd$,
where $d$ indexes the spatial location in $\bm{y}_j$;
$q$ is the index considering the width, height and channel in the map.
The zoom network on level $m$ is trained using the cross-entropy loss  plus a regression loss: 
\begin{align}
\mathcal{L}^{(m)}\big( \mathcal{B} | \bm{w} \big)
=  \sum_i  - \log p_{i k^{*}} +  \delta (l_{ik^*})  \big \| \bm{t}_i - \bm{t}_i^{*}  \big \|^2,
\label{loss}
\end{align}
where $\delta(\cdot)$ is the indicator function.
$\bm{p}_i= \{ p_{ik} \}  \in \mathbb{R}^{K}$ 
is the estimated probability with $K$ being the total number of classes. In our task, $k=0,1$.
$\bm{l}_i = \{ l_{ik} \}$ denotes the label vector with its correct label index $k^*$ being 1 and others being 0.
$\bm{t}_i, \bm{t}^{*}_i \in \mathbb{R}^4$ indicates the estimated and  ground truth regression offset \citep{fast_rcnn}, respectively.
The total loss of the zoom network is optimized as the summed loss across all levels:
$\mathcal{L}_{zoom\_net} = \sum_{m} \mathcal{L}^{(m)} \big(  \mathcal{B}   \{  \bm{y}_j  (i)  \} | \bm{w} \big).$
Note that the number of samples $\mathcal{B}$ in a mini-batch  varies in each level, since there are way more  anchors in lower depth than those in higher one.

The upper gradient assumed to be existent in Eqn.(\ref{MAD_grad}) and (\ref{MAD_2}) is therefore derived as:
\begin{align}
\nabla_{\bm{y}_j} \mathcal{L} = & ~ \frac{\partial \mathcal{L}}{ \partial y_{jd}} \triangleq  \frac{\partial \mathcal{L}}{ \partial y_{q}} =  \nonumber\\
= & ~ \sum_k  \frac{\partial \mathcal{L}}{  \partial p_{k}} \frac{ \partial  p_{k}}{\partial y_q}
+ \delta(l_{k^*}) \sum_r  \frac{\partial \mathcal{L}}{ \partial  t_{r}} \frac{  \partial t_{r}}{\partial y_q}, \nonumber \\
=& ~ \sum_k (p_k - l_k) w_{qk} + \delta(l_{k^*})  \sum_r (t_r - t^*_r) w_{qr},
\end{align}
where we drop notations of level $m$ and sample $i$ for brevity. $w_{qk}$ and $w_{qr}$ are the `abstract' weights from input $y_q$ to the loss layer $\mathcal{L}$. Note that there could be several different layers between the attention maps and the final RPN loss.

Figure \ref{fig:featMap} visualizes the effect of using MAD unit on acquiring more accurate neuron activations among feature maps. Take the feature maps at level $m=2$ (for medium objects) as an illustration, we can see the low-level features contain many local details and might be noisy for detection; while raw high-level features in $\bm{H}^2$ highlight some region around the helmets, there are still many disturbing areas. A naive combination of high-level and low-level maps by convolution can effectively remove unnecessary background details and magnify objects on medium size.
However, we can observe from (h) that by enforcing a MAD unit to actively look for neuron attention among feature channels, the outcome could have clear and strong signal as to where to find potential objects, which could benefit the classification and regression in the RPN layer in a great deal. 

\subsection{Training Strategies and Anchor Design}\label{sec:anchor-templates}

There are several remarks regarding the training of the zoom network.
First,
\textit{adjust input scale dynamically}, known as  \texttt{dyTrainScale} in Table \ref{tab:design}.
Each sample is resized to the extent where at least one of 
the ground truth boxes
is covered by the medium anchors. 
This ensures there are always positive samples in each batch and the model can relatively see a fixed range of anchor sizes - achieving multi-scale training at low computational cost.
Second,
\textit{control the number of negative samples}, denoted as \texttt{equilibrium}.
In preliminary experiments, 
we find  that the training loss converges slowly if we fill in the rest of a batch with negative samples. 
This will 
cause the unbalance of training data when the number of positive samples is small. 
Thus, we strict the number of negative samples to be at most twice the number of the positive.
Third,
\textit{add an additional gray class}, indicated as \texttt{grayCls}. Adding an additional gray label into the training stage  better separates the positive from the negative. For the positive class, IoU threshold is above 0.6; for the gray class, IoU ranges from 0.35 to 0.55; and for the negative, the criterion is below 0.25.
The number of gray samples is set to be half of the total number of positive and negative ones. 
Note that these  strategies are not mentioned in the original region proposal networks \citep{faster_rcnn}.

\begin{figure}[t]
	\begin{center}
		\includegraphics[width=.49\textwidth]{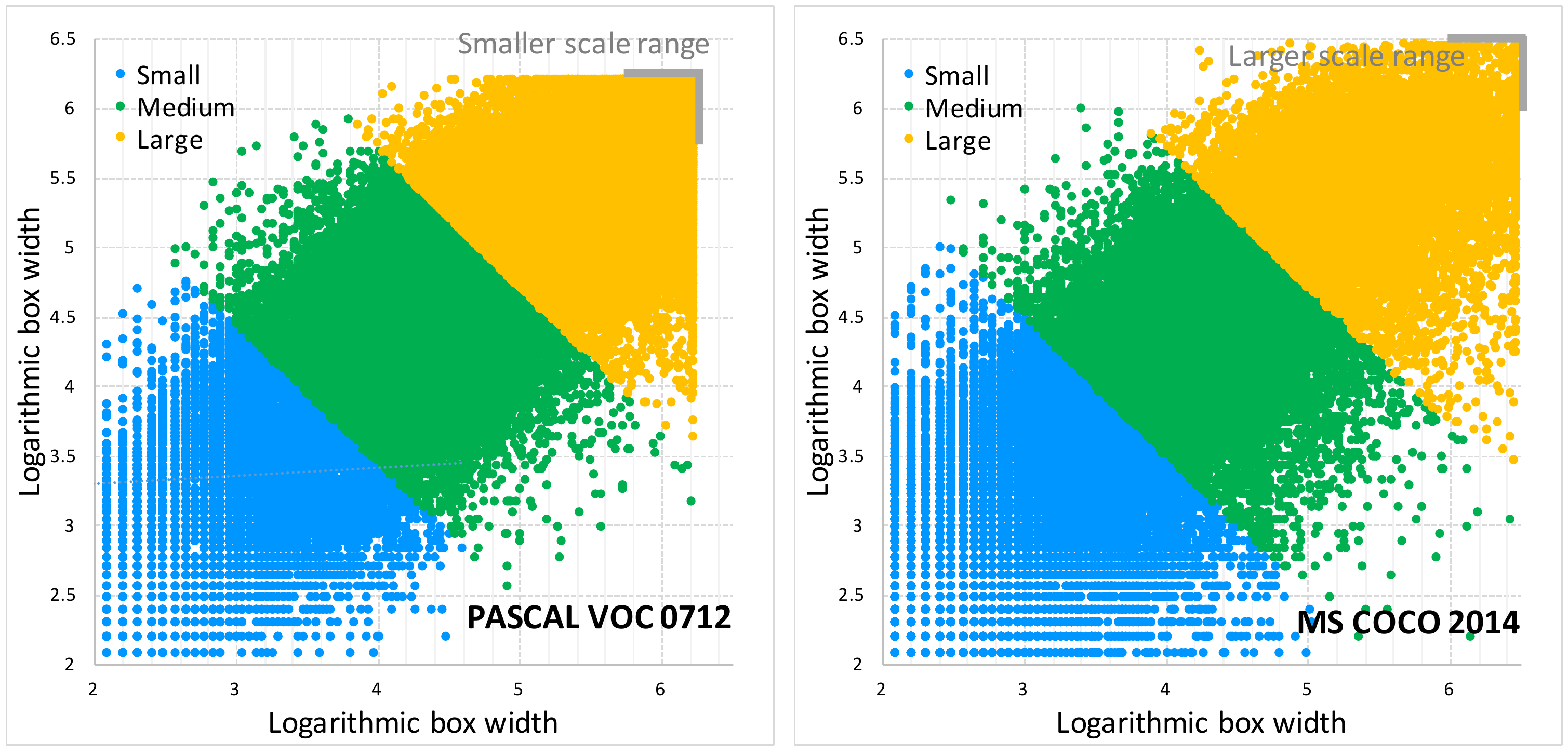}
	\end{center}
	\caption{Anchor distribution in logarithmic width and height. Left: PASCAL VOC 0712 \texttt{trainval} data; right: MS COCO 2014 \texttt{test} data. Region proposals share similar properties across datasets; on COCO the anchors vary slightly more in terms of size and aspect ratio.}
	\label{fig:dist}
\end{figure}

As stated in Section \ref{Sec:intro}, we place anchors of  various sizes  at different depth in the  network.  Figure \ref{fig:dist} depicts the anchor distribution in PASCAL VOC  and COCO.
The training and test sets share similar data properties \textit{within} and \textit{across} dataset.
There are 30 anchors in total and each depth holds 10 box templates.
Since the scale problem is handled by the various resolutions in the feature map, we set the number of different aspect ratios (which is 5) more than the number of scales (which is 2 in our setting) in each level.
We manually group the anchors into three clusters based on the scale distribution across datasets.
Specifically,
the scale of anchors in each level are $\{16, 32\}$, $\{64, 128\}$, $\{256, 512\}$, respectively; and the aspect ratio set is $0.25, 0.5, 0.75, 1, 2, 4$. Note that these anchor templates are unanimously set across all datasets to be evaluated. 

In SSD \citep{ssd}, the design of anchors is heuristic and exhaustive: it places too many templates in  (relative) size from 0.2 to 0.9  of the fixed input image on six locations. In our work, we bear in mind the computational cost and design the anchor size carefully: we divide the anchor  space into three domains and vary the input image size such that there will always be objects of size around 128 (the medium size); the anchors are placed only at three locations (depth) in the network, compared to the six prediction modules in SSD. The size of feature maps in each branch of our method is exactly halved to the previous branch, thus covering a vast majority of  anchor space at a lower computation budget.

\subsection{Zoom Network Prediction}\label{sec:prediction}

During inference, we merge proposal results from all the three levels. Prior to merging,
an initial NMS \citep{objectness} process \textcolor{black}{(with top boxes 2000 and IoU threshold 0.5)} is launched to first filter out low confidence and redundant boxes.
Since the raw  scores  at each level is not comparable, we tune a bias term for each level to add on the raw scores for better combination among results. Note that these thresholds are cross validated on a smaller set different from the one used for evaluation.
The second NMS (with top boxes 300) is conducted in a multi-threshold manner as does in \citep{attractioNet} to automatically set the optimal IoU threshold under different number of proposals to be evaluated.
Since the scale varies dynamically during training, we also forward the network in several image scales, 
ranging from 1000 to 250 with interval 250.

\section{Zoom Network for Object Detection}\label{sec:zoom-network-with-recursive-training-for-object-detection}

We now verify our proposed method in the context of object detection and embed the MAD unit, which is for selecting feature maps from various branches in the network, with region proposal generation in an end-to-end learning system.
\begin{figure*}[t]
	\begin{center}
		\includegraphics[width=.8\linewidth]{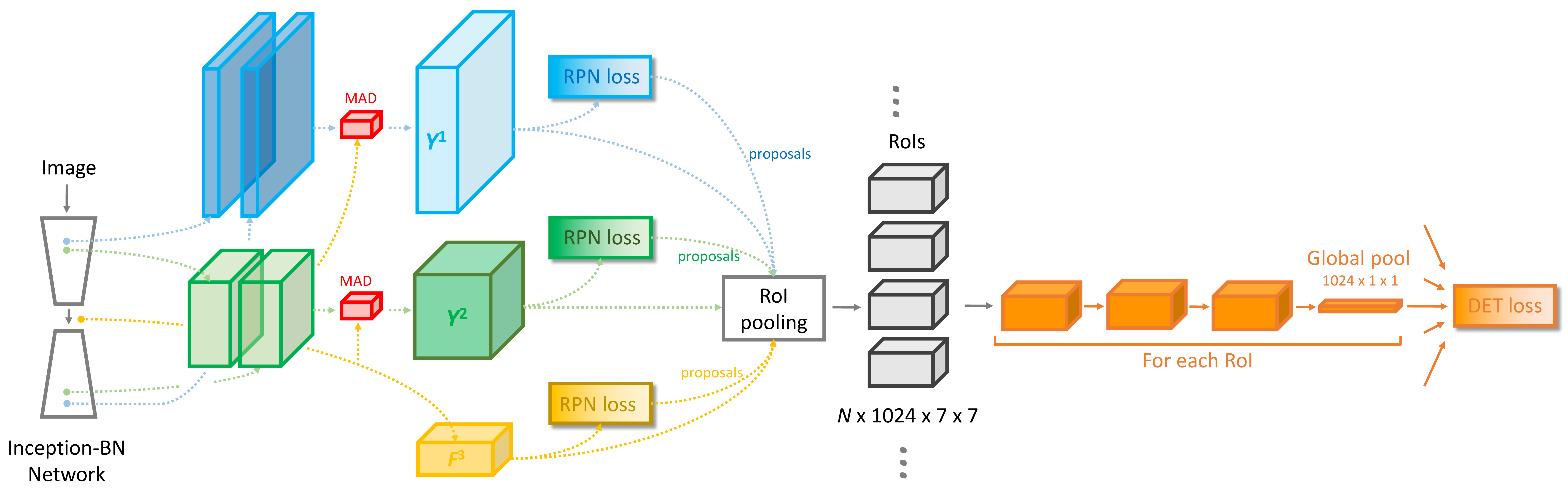}
	\end{center}
	\vspace{-.3cm}
	\caption{
		The complete pipeline of ZIP and MAD algorithm for region proposal and object detection. The system can be trained end-to-end.
		Based on the proposals generated from the zoom network, we conduct a RoI-pooling on each level after the MAD unit. 
		$N$ is the number of RoIs in one mini-batch. Note that the detection follow-up layers (marked in orange after RoI-pooling) 
		share the \textit{same} parameter across different levels, which is different from Fig. \ref{fig:pipeline} (b).
	}
	\label{fig:pipeline_det}
\end{figure*}
Figure \ref{fig:pipeline_det} depicts the detection follow-up pipeline. Since the channel number of attended maps on different level varies (640, 1216 and 1024 from level 1 to 3, respectively), we have an additional convolution layer for each level, which transfers the RoI-pooled feature maps to have the same number of channels after the RoI-pooling (1024 in our setting, not shown in the figure). The network architecture for the detection follow-up layers consists of 9 blocks, corresponding to the higher modules in ResNet-101 model \citep{resNet}, \textit{i.e.}, \texttt{res4b15} to \texttt{res4b20}, \texttt{res5a} to \texttt{res5c}.
The detection loss, denoted as $\mathcal{D}\big( \mathcal{B} | \bm{w} \big)$, resembles Eqn. (\ref{loss}), where the summation traverses over the number of RoIs and the class label $k$ ranges from 0 (background) to the number of total object classes in one dataset.
Note that we have the parameter sharing strategy among RoI-pooled features from various levels, which allows us to 
have more  follow-up layers  afterwards (crucial to the classification confidence score) in the detection pipeline.

\textbf{MAD extension and feature boosting.} To further generalize the MAD unit and boost the detection performance, we have the following two extensions.
In the original formulation (\ref{MAD_basic}), the length of the MAD unit is exactly twice as the number of the incoming feature maps, since we merge the two sources from high-level and low-level, respectively. The number of incoming feature maps can be various. Therefore, we extend the formulation of MAD to:
\begin{equation}
\bm{\mu}^{(m)} = \texttt{MAD}(\bm{F}^3, m) \in \mathbb{R}^{ 1 \times 1 \times \lambda C^{(m)}}, \label{MAD_extend}
\end{equation}
where $\lambda$ could be $2,4,\dots$. The combination of map sources could be various and manually designed. 
Second, we modify the detection confidence score from the neural network output $p$ to the classifier decision of the boosting algorithm. Specifically, for each RoI, we extract features from the follow-up detection blocks (marked in red in Figure \ref{fig:pipeline_det}
), concatenate them and feed into the Adaboost classifier, in favor of leveraging different aspects of features in the network and forging into a stronger classifier for detection. Details for the configuration of these two extensions are 
provided in Section \ref{experiments}. 


%% file: experiments.tex
\section{Experiments}\label{experiments}

\noindent We evaluate the effectiveness of our algorithm  on generating 
region proposals and performing generic object detection. To this end, we conduct ablative evaluation of  individual components in our approach on PASCAL VOC \citep{pascal} and compare results with state-of-the-arts on two
challenging 
datasets, ILSVRC DET 2014 \citep{imagenet_conf} and Microsoft COCO \citep{coco}.

\subsection{Datasets and Setup}

The PASCAL VOC dataset \citep{pascal} is designed for object detection and contains 20  object classes. To fast verify our algorithm's design, we conduct experiments on this dataset, where we use the training and validation set in 2007 and 2012 (16551 images) during training and evaluate on the 2007 test set (4952 images).
The ILSVRC DET 2014 dataset \citep{imagenet_conf} is a subset of the whole ImageNet database and consists of more than 170,000 training and 20,000 validation images. Since some training images has only one object with simple background, which has a  distribution discretion with 
the validation set, we follow the practice of  \citep{fast_rcnn} and split the validation set into two parts. The training set is the \texttt{train\_14} with 44878 images and \texttt{val1} with 9205 images. 
We use \texttt{val2} as the validation set for evaluation.
The Microsoft COCO 2014 dataset \citep{coco} contains 82,783 training and 40,504 validation images, where most images have various shapes surrounded by complex scenes.
We use all the training images, without any data augmentation, to learn our model, and follow \citep{deepMask} to evaluate on the first 5000 validation images (denoted as \texttt{val\_5k}).

\textbf{Implementation details.} We reimplement an Inception BN \citep{bn} model on the ImageNet classification dataset,
which could achieve around 94\% top-5 classification accuracy.
The zoom network is finetuned using the pretrained model and the zoom-in part is also initialized by coping the weights from its mirror layers.
The convolution layers in MAD unit is trained from scratch.
The base learning rate is set to 0.0001 with a 50\% drop every 7,000 iterations. The momentum and weight decay is set to be 0.9 and 0.0005, respectively. The maximum training iteration are roughly 8 epochs for each dataset. We utilize stochastic gradient descent for optimization.
The mini-batch size $\mathcal{B}$ during each iteration is set to be 300 with each class having at most 100 samples.
Note that during training, there are always some cases where the actual batch size is lower than the pre-setting due to the sample equilibrium scheme stated in Section \ref{sec:anchor-templates}. The network architecture in the following experiments are derived (zoom-out part) and mirrored (zoom-in part) from the Inception-BN model, which is explicitly stated in Section \ref{sec:network-architecture}.

\textbf{Evaluation metric.} We use recall (correctly retrieved ground truth boxes over all annotations) under different IoU thresholds and number of proposals as metric to evaluate region proposals; and precision (correctly identified boxes over all predictions) to evaluate object detection. The mean value of recall and precision from IoU 0.5 to 0.95 is known as average recall (AR) and precision (AP). AR summarizes the general proposal performance and is shown to correlate with the average precision (AP) performance of a detector better than other metrics \citep{Hosang2015Pami}.
Moreover,
we compute AR and AP of different sizes of objects to further investigate on a specific scale of targets. 
Following COCO-style evaluation, we denote three types of instance size, @small ($\alpha < 32^2 $), @medium ($ 32^2 \le \alpha < 96^2 $) and @large ($ 96^2 \le \alpha$), where $\alpha$ is the area of an object. 

\begin{table}
	\caption{Ablation  on the  network structure. AR on PASCAL VOC is reported. We use 30 anchors and treat training as a two-class problem. `\texttt{sp}' means splitting the anchors. Four out of five cases below (a, b, d, e) are illustrated in Figure \ref{fig:structure}.
	}\label{tab:net_structure}
	\vspace{-.3cm}
	\begin{center}
		\footnotesize{
			\begin{tabular}{l c c c c }
\toprule
Structure & \scriptsize AR@100 & \scriptsize AR@S & \scriptsize AR@M & \scriptsize AR@L  \\
\midrule
(a) Zoom-out, \scriptsize{F-RCNN} &  62.31  &  40.77&  57.97&  71.05 \\
(b) Zoom-out\_\texttt{sp} & 65.98 & 43.81& 60.35& 73.87\\
\textcolor{black}{(c) Zoom-out\_\texttt{sp}, {all} } & 66.02 & 44.52 & 61.77 & 74.13 \\ 
(d) ZIP & \textbf{ 68.51} & \textbf{49.07} & \textbf{62.93} & \textbf{75.64} \\
(e) Deeper Zoom-out\_\texttt{sp} & 67.21 & 47.92& 61.53& 74.02\\
\bottomrule
			\end{tabular}
		}
	\end{center}
\end{table}

\subsection{Ablative Evaluation for Region Proposals}\label{sec:ablation-study}

All the experiments in this subsection 
are conducted 
on the PASCAL VOC dataset for the region proposal task. 

\textbf{Network structure design.} 
Table \ref{tab:net_structure} reports AR for different network design. All images are fed into the network with a fixed size on shorter dimension of 600. The anchor design is mentioned in Section \ref{sec:anchor-templates}. Other settings are by default as that in  \citep{faster_rcnn}. There are some observations:
(a) if we employ a zoom out structure and place all anchors at the last feature map, the recall is 62.31\%, which is adopted by most popular detectors, \textit{e.g.}, Faster-RCNN \citep{faster_rcnn};
(b) then we split the anchors into three groups and insert them at layer \texttt{icp\_3b}, \texttt{icp\_4d} and \texttt{icp\_5b}.
Such a  modification will increase  recall to 65.98\%. 
It indicates that extracting object proposals at different depth helps;
\textcolor{black}{
(c) if we consider another comparison where all anchors are placed at each desired location in the network, \textit{i.e.}, \texttt{icp\_3b}, \texttt{icp\_4d} and \texttt{icp\_5b}, the average recall is 66.12. The result is better than (b) because it places all anchors, no matter big or small, densely into the network - boosting the search space of object boxes. However, the computational cost of (c) is around 3.5 times than (b) due to the heavy burden on dense anchors in lower layers;}
(d) the zoom-out-and-in design (ZIP) further increases performance of  recall to 68.51\%. 
\textcolor{black}{
ZIP increases the network depth and has about 40 layers; 
(e) by simply stacking layers of the network to 40 layers via a zoom-out design, we do not witness an obvious increase compared with ZIP, which verifies that the gain in our method does not come from depth or parameter increase of the model.}
%
Compared with the baseline zoom-out structure, ZIP achieves a larger improvement of AR on small objects (8.3\%) than medium-sized objects (4.96\%)
and large objects (4.59\%). Such a zoom-out-and-in structure could enhance the quality of proposals in all ranges of scales.

\begin{table}
	\caption{Ablation study on MAD unit and training strategies. AR on PASCAL VOC is reported. 
		We use the zoom-out-and-in network structure. 
		Baseline model corresponds to the (d) setting in Table \ref{tab:net_structure}.
		\texttt{all} means adopting all three training strategies. \texttt{ER} vector denotes the Excitation Recalibration  idea implemented in SENet \citep{SENet}. See context in the paper for setting details.
	}
	\label{tab:design}
	\vspace{-.3cm}
	\begin{center}
		\footnotesize{
			\begin{tabular}{l c  c c c }
				\toprule
				Strategy &  AR@100 & AR@S & AR@M & AR@L \\
				\midrule
				Baseline & 68.51 & 49.07 & 62.93 & 75.64\\
				\midrule
				\texttt{dyTrainScale} &  72.58  & 53.41& 66.34& 78.83\\
				\texttt{equilibrium} & 69.85 & 50.49 &63.55 & 76.88 \\
				\texttt{grayCls} & 69.77  & 50.21 & 63.19 & 77.04 \\
				\midrule
				ZIP + \texttt{all} &  74.22 &  54.39 & 68.47 &  81.53\\
				ZIP + \texttt{all} + MAD & \textbf{76.51} &  57.28  & 70.05 & 84.21 \\  
				\midrule
				\textcolor{black}{\texttt{spatial} MAD } & 76.50& 57.11 & 70.13& 84.09 \\
				\textcolor{black}{\texttt{ER} vector }& 71.87 & 51.84& 65.49& 76.21\\
				\bottomrule
			\end{tabular}
		}
	\end{center}
\end{table}

\textbf{Training strategies.} 
Table \ref{tab:design} shows 
the effect of MAD unit and different training strategies stated in Section \ref{sec:sop-algorithm}. 
Starting from the baseline, we add each training component individually and investigate their contribution to AR. It is observed that the dynamic alternation of input scale weighs more among these strategies. The overall AR improvement of merging all strategies could obtain a recall of 74.22\%. This result is the outcome of imposing a simple convolution kernel to combine features from two streams (qualitatively visualized in Figure \ref{fig:featMap}(g)).
Furthermore,
the MAD unit can actively attend important neuron activations and we verify its effectiveness via the last experiment: enhancing AR to 76.51\%.

\textcolor{black}{\textbf{Other alternatives to MAD.} A natural thought could be to skip the global pooling after the convolution operation at the output of \texttt{icp5}. The MAD vector $\bm{\mu}^{(m)}$ could be spatially sized and proportional to each input image and level depth. In help of deconvolution, we can align spatially the MAD vector with feature maps to be weighted. In this manner, each pixel, instead of all pixels on a channel, could be evaluated  in terms of  importance towards the final loss. Such a design is denoted as `\texttt{spatial} MAD' in Table \ref{tab:design}. We can see the performance is similar compared to the global pooling version. Since it could increase the computational cost by adding more parameters in the model, we do not resort to this alternative.
Such a result indicates that for weighing the importance of feature maps, the learned information along the channel dimension is more important, or at least enough, than the information learned along both the channel and spatial dimensions.
	We also try the excitation recalibration idea to replace MAD. Specifically, $\bm{Y}^{(m)}$ is generated from the squeeze-and-excitation unit alone. For $m=1,2$, the input is the concatenation of $\bm{F}^{(m)}$ and $\bm{H}^{(m)}$; for level $m=3$, the input is $\bm{F}^{(3)}$. As is shown in the last case of Table \ref{tab:design}, the replacement is  inferior to ours (71.87 vs 76.51). This is probably because that the excitation vector derives from  information directly in the preceding layer - an inherent design in \citep{SENet}; whilst our MAD comes from  higher-level summation, leveraging the guidance of high-level, region-based semantics to merge local details in lower layers.
}

\begin{figure*}
	\begin{center}
		\includegraphics[width=.5\textwidth]{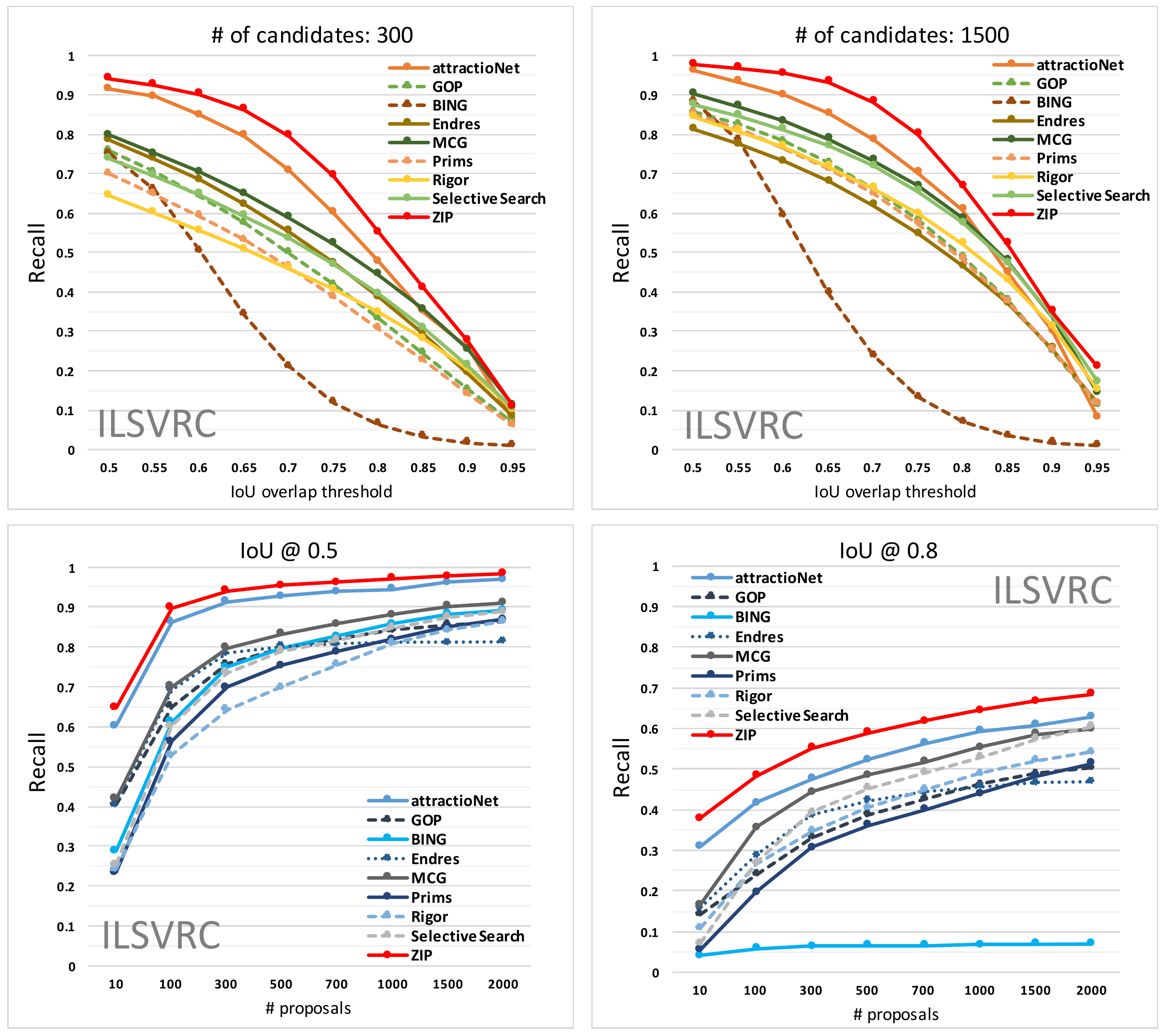}\includegraphics[width=.5\textwidth]{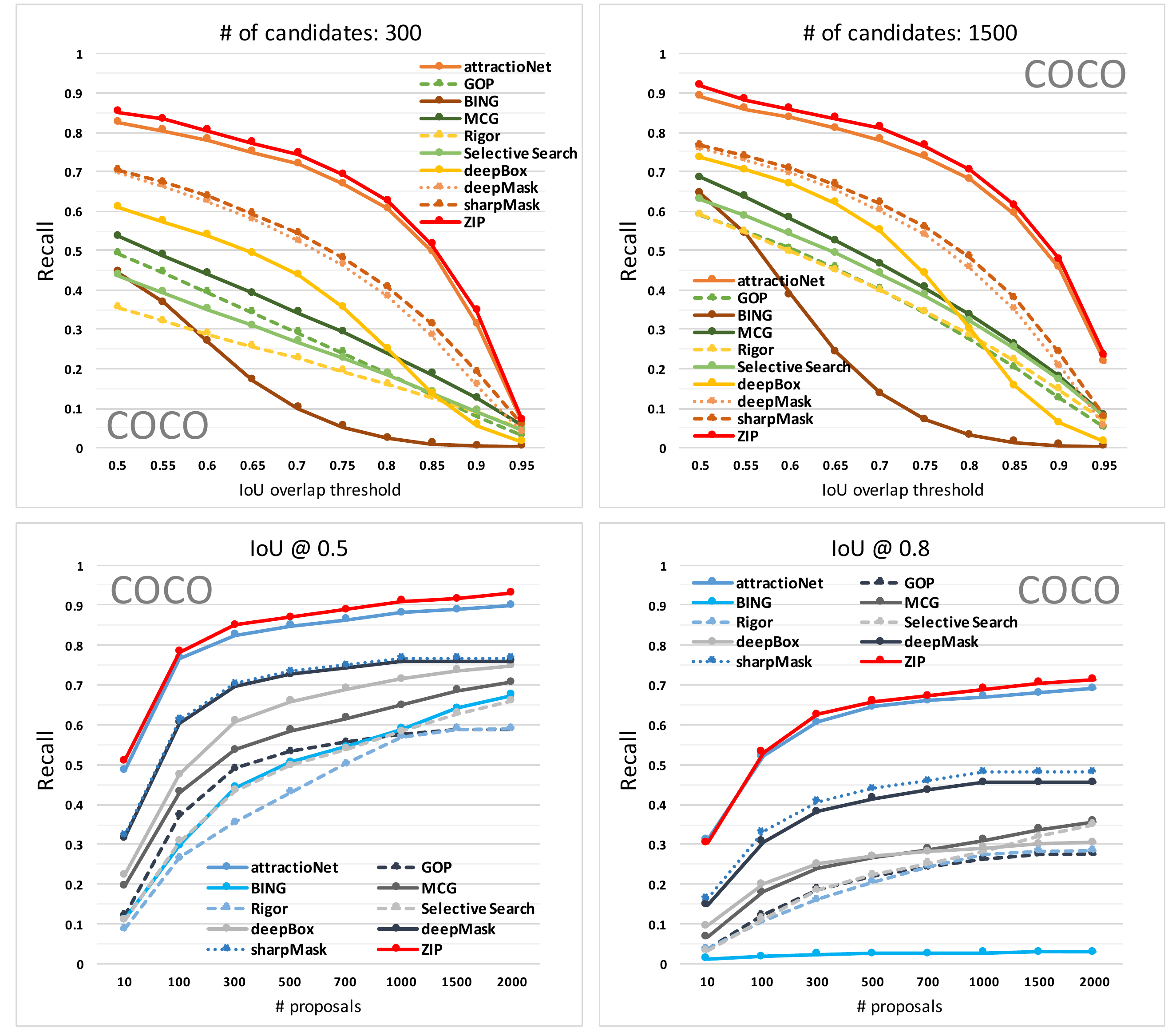}
	\end{center}
	\vspace{-.2cm}
	\caption{Recall at  different IoU thresholds (0.5, 0.8) and number of proposals (300, 1500). Left: ILSVRC DET 2014 \texttt{val2}. Right: MS COCO 2014 \texttt{val\_5k}.
		For method abbreviations, please refer to the context. 
	}
	\label{fig:recall_detail}
\end{figure*}

\textbf{Runtime analysis.} 
During training the image is resized to a much smaller size based on the medium anchor size ($64-128$) and thus the total training time is one third less than that without such a constraint. For inference, the runtime per image is 1.13s (Titan X, code partially optimized), compared with AttractioNet \citep{attractioNet} 1.63s and DeepMask \citep{deepMask} 1.59s.

In the following experiments, we denote our method as `ZIP' and by default it includes  MAD  if not specified.

\subsection{Average Recall  for Region Proposals}

Figure~\ref{fig:recall_detail} illustrates recall under different IoU thresholds and number of proposals.
Our algorithm is superior than or on par with previous state-of-the-arts, including:
BING \citep{bing},
EdgeBox \citep{edgebox},
GOP \citep{gop},
Selective Search \citep{selective_search},
MCG \citep{MCG},
Endres \citep{cat_ind_obj_prop},
Prims \citep{prime},
Rigor \citep{rigor},
Faster RCNN \citep{faster_rcnn},
AttractioNet \citep{attractioNet},
DeepBox \citep{deep_box},
CoGen \citep{co_generate},
DeepMask \citep{deepMask},
SharpMask \citep{SharpMask},
and FPN \citep{pyramid_feature_learn}. 
Table \ref{tab:grand_ilsvrc} 
reports the average recall vs. the number of proposals (from 10 to 1000) and the size of objects on ILSVRC.
Our method 
performs better on identifying different size of objects.
This proves that the zoom-out-and-in structure with a MAD unit is effective.
For MS COCO, similar phenomenon of AR is also observed
via Table \ref{tab:grand_coco}.
We  include results from the latest work \citep{pyramid_feature_learn} which shares similar spirit of a conv-deconv structure and combing feature maps from different sources. The main distinction from ours to \citep{pyramid_feature_learn} is the introduction of MAD unit to dynamically decide the attention weights of feature channels based on the context of the input image.


Compared with Faster RCNN \citep{faster_rcnn} 
(which also employs CNN, has a zoom-out structure with all anchors placed at the final layer and thus serves as as our baseline), ZIP has achieved a very large improvement over the baseline on small objects (21.8\%) and medium objects (22.4\%). The performance improvement for large objects is also satisfying (8.4\%).
AttractioNet \citep{attractioNet} also has large improvements on small and medium object proposals, since it employs the recursive regression and active box generation. These techniques are also beneficial for detecting small objects.

\begin{table*}[t]
	\caption{Average recall (AR) analysis on ILSVRC \texttt{val2}. The AR for small, medium and large objects are computed for 100 proposals. The top two results in each metric are in {\textbf{bold}} and {\textit{italic}}, respectively. 
	}
	\vspace{-.3cm}
	\begin{center}
		\footnotesize{
			\begin{tabular}{l c c c c c c c}
				\toprule
				\textbf{ILSVRC DET 2014} & AR@10 & AR@100 & AR@500 & AR@1000 & AR@Small & AR@Medium & AR@Large \\
				\midrule
				BING 
				& 0.114 & 0.226 & 0.287& 0.307 & 0.000 & 0.064 & 0.340 \\
				EdgeBox 
				&  0.188 & 0.387 &  0.512 &0.555&0.021 &0.156 & 0.559  \\
				GOP 
				& 0.208 & 0.349 & 0.486 & 0.545 & 0.022 & 0.185 & 0.482 \\ 
				Selective Search 
				& 0.118 & 0.350 & 0.522 & 0.588 & 0.006 & 0.103 & 0.526\\
				MCG 
				& 0.229 & 0.435 & 0.553 & 0.609 & 0.050 & 0.215 & 0.604   \\
				Endres 
				& 0.221 & 0.393 & 0.508 & 0.531 & 0.029 & 0.209 & 0.543  \\
				Prims 
				&  0.101 & 0.296 & 0.456 & 0.523 & 0.006 & 0.077 & 0.449 \\
				Rigor 
				& 0.139 & 0.325 & 0.463 &0.551 & 0.027 & 0.092 & 0.485 \\
				Faster RCNN 
				& 
				{}{0.356} & 0.475 & 0.532 & {}{0.560} & 0.217 & {}{0.407} & 0.571
				\\
				AttractioNet 
				& {}{\textit{0.412}} & {}{\textit{0.618}} &  \textit{0.672} & \textit{0.748} &  \textit{0.428} & \textit{0.615} & \textit{0.623} \\
				\midrule
				Zoom Network (ZIP) & \textbf{0.420} & \textbf{0.635} & \textbf{0.688} & \textbf{0.761} & 
				\textbf{0.435} & \textbf{0.631} & \textbf{0.655} \\
				\bottomrule
			\end{tabular}
		}
	\end{center}
	\label{tab:grand_ilsvrc}
\end{table*}

\begin{table*}[t]
	\caption{Average recall (AR) analysis on COCO \texttt{val\_5k}. The AR for small, medium and large objects are computed for 100 proposals. The top two results in each metric are in {\textbf{bold}} and {\textit{italic}}, respectively.  
	}
	\vspace{-.5cm}
	\begin{center}
		\footnotesize{
			\begin{tabular}{l c c c c c c c}
				\toprule
				\textbf{MS COCO 2014} & AR@10 & AR@100 & AR@500 & AR@1000 & AR@Small & AR@Medium & AR@Large  \\
				\midrule
				BING 
				& 0.042 & 0.100 & 0.164 & 0.189 & 0.000 & 0.026 & 0.269 \\
				EdgeBox 
				&  0.074 & 0.178 & 0.285 & 0.338 &0.009 & 0.086 &0.423\\
				GOP 
				& 0.058 & 0.187 & 0.297 & 0.339 & 0.007 & 0.141 & 0.401 \\
				Selective Search 
				& 0.052 & 0.163 & 0.287 & 0.351 & 0.003 &0.063 &0.407\\		
				MCG 
				&  0.098 & 0.240 & 0.342 & 0.387 & 0.036 & 0.173 & 0.497 \\
				Endres 
				& 0.097 & 0.219 & 0.336 & 0.365 &0.013 & 0.164 &0.466\\
				DeepBox 
				& 0.127 & 0.270 & 0.376 &0.410 &0.043  & 0.239 & 0.511 \\
				CoGen 
				&0.189 & 0.366 & -  & 0.492 & 0.107 & 0.449 & 0.686 \\
				DeepMask 
				& 0.183 & 0.367 & 0.470 & 0.504 & 0.065 & 0.454 & 0.555 \\
				SharpMask 
				& {}{0.196} & 0.385 & 0.489& 0.524 & 0.068 & {}{0.472} &{}{0.587}\\
				FPN 
				& - & 0.440 & - & 0.563 & - & - & -\\
				AttractioNet 
				&\textit{0.328} & {}{\textit{0.533}} & {\textit{0.601}}& {}{\textit{0.662}} & {}{\textit{0.315}} & {}{\textit{0.622}} & {}{\textit{0.777}}  \\
				\midrule
				Zoom Network (ZIP) &   \textbf{0.335} & \textbf{0.539} & \textbf{0.612} & \textbf{0.670} & \textbf{0.319} & \textbf{0.630} & \textbf{0.785} \\
				\bottomrule
			\end{tabular}
		}
	\end{center}
	\label{tab:grand_coco}
\end{table*}

\begin{table}[t]
	\caption{ Generalization ability of ZIP when the training and test set are from different datasets. 
		The test set is fixed as ILSVRC. When the training set is changed from ILSVRC to COCO, AR@10 only drops slightly.	
		Definition of seen and unseen categories are the same with AttractioNet \citep{attractioNet}. 
	} 	 	
	\vspace{-.4cm}
	\begin{center}
		\footnotesize{
			\begin{tabular}{l c c c c c c }
				\toprule
				Train Data & All & Seen & Unseen \\
				\midrule
				ZIP on ImageNet & 42.0 & 48.5 &  33.3\\
				ZIP on COCO   &  41.9&   48.23 &  31.7\\
								AttractioNet on COCO & 41.2 & 47.41 & 29.9 \\
								MCG on  COCO & 21.9 & 22.8 & 20.5 \\
				\bottomrule
			\end{tabular}
		}
	\end{center}
	\vspace{-.5cm}
	\label{tab:generalize}
\end{table} 

\textbf{Generalization for unseen categories.}
In order to evaluate the generalization capability, we show   the average recall @10 boxes in Table \ref{tab:generalize}, where our ZIP model is
trained on COCO and tested on ILSVRC \texttt{val2} set. Although COCO has fewer classes, which is 80, the scales and aspect ratios of objects varies in great extent  and thus the trained ZIP model can handle most categories on ILSVRC DET dataset, which has 200 classes.

\subsection{Ablative Evaluation for Object Detection}\label{sec:ablation-study-detection}

All the experiments in this subsection 
are conducted 
on the PASCAL VOC dataset for  object detection. The backbone structure (inception-BN) for detection is the same as that in region proposal. \textcolor{black}{We have conducted experiments based on the VGG-16 model and the performance is slightly inferior (76.4\%, \textit{c.f.} 76.8\% denoted as `ZIP MAD' in Table \ref{tab:voc07_all}). Hence we employ the inception model thereafter. }

Table \ref{tab:voc07_all}
reports the detailed ablation investigation on the object detection task. For reference, we also list the results from popular methods, RFCN \citep{rfcn}, SSD \citep{ssd}, Faster-RCNN \citep{faster_rcnn}, in the first three rows.
We can see the zoom-in architecture alone could enhance the performance in a great deal (\textit{c.f.}, zoom-out and ZIP baseline, around 2\%). The MAD unit could further improve the result by leveraging feature weights among different channels in the network.  \textcolor{black}{Invoking the boosting method alone into the pipeline (denoted as `ZIP MAD \texttt{B}', using all features in the nine blocks) could enhance the detection performance to some extent.} Empirically we find the value of $\lambda$ in Eqn. \ref{MAD_extend} to be 4 best fits the task, which is noted as `ZIP MAD \texttt{E}' in the table. At last, after incorporating both the boosting and extension schemes, we have the final result on PASCAL to be 79.8\% mAP; such an improvement bears from the zoom-in structure, the auto-selected MAD unit and the stronger classifier from boosting.
%


\newcolumntype{x}[1]{>{\centering}p{#1pt}}
\newcolumntype{y}{>{\centering}p{16pt}}
\newcommand{\hl}[1]{{#1}}
\newcommand{\ct}[1]{\fontsize{7pt}{1pt}\selectfont{#1}}
\renewcommand{\arraystretch}{1.2}
\setlength{\tabcolsep}{1.5pt}
\begin{table*}[t]
	\caption{Ablation study for object detection on the PASCAL VOC 2007 test set. The training data comes from both 2007 and 2012 training and validation set. `Zoom-out' indicates that the network structure is the standard convolution type without up-sampling. \texttt{E} means MAD extension; \texttt{B} means boosting and \texttt{EB} denotes the combination.}
	\label{tab:voc07_all}
	\vspace{-.2cm}
	\centering
	\footnotesize
	\resizebox{\linewidth}{!}{
		\begin{tabular}{l|x{30}|x{20}|yyyyyyyyyyyyyyyyyyyc}
			\toprule
			method & data & mAP & \ct{areo} & \ct{bike} & \ct{bird} & \ct{boat} & \ct{bottle} & \ct{bus} & \ct{car} & \ct{cat} & \ct{chair} & \ct{cow} & \ct{table} & \ct{dog} & \ct{horse} & \ct{mbike} & \ct{person} & \ct{plant} & \ct{sheep} & \ct{sofa} & \ct{train} & \ct{tv} \\
			\midrule
			\footnotesize Faster R-CNN & 07+12 & 76.4 & 79.8 & 80.7 & 76.2 & 68.3 & 55.9 & 85.1 & 85.3 & {89.8} & 56.7 & 87.8 & 69.4 & 88.3 & 88.9 & 80.9 & 78.4 & 41.7 & 78.6 & 79.8 & 85.3 & 72.0 \\
			\footnotesize SSD512 & 07+12 & 76.8 &82.4 &
			84.7&78.4&73.8&53.2&86.2&87.5&86.0&57.8&83.1&70.2&84.9&85.2&83.9&79.7&50.3&77.9&73.9&82.5&75.3 \\
			\footnotesize R-FCN  & 07+12 & 79.5 &82.5 &83.7 &80.3 &69.0 &69.2 &87.5 &88.4 &88.4 &65.4 &87.3 &72.1 &87.9 &88.3 &81.3 &79.8 &54.1 &79.6 &78.8 &87.1 &79.5\\
			\midrule
				Zoom-out  & 07+12  &71.3  & 72.1 & 77.3 & 73.4 & 62.8 & 50.1 & 79.6 & 81.2 & 82.7 & 49.6 & 82.6 & 65.4 & 83.7 & 81.4 & 73.5 & 75.4 & 38.3 & 74.1 & 73.6 & 81.1 & 68.5 \\
			ZIP baseline & 07+12  &73.1  & 73.8 & 79.2 & 75.8 & 64.2 & 53.7 & 81.3 & 83.0   & 85.2 & 51.4 & 85.7 & 66.2 & 84.4 & 83.7 & 75.1 & 76.8 & 40.5 & 76.2 & 75.4 & 83.1 & 67.9 \\
			 ZIP MAD & 07+12 & 76.8 & 74.1 & 80.1 & 76.0   & 65.7 & 55.4 & 83.5 & 85.3 & 87.5 & 54.2 & 87.3 & 68.6 & 87.4 & 85.2 & 77.2 & 77.7 & 40.9 & 77.8 & 77.5 & 84.6 & 70.3 \\
			 \textcolor{black}{
ZIP MAD \texttt{B}} & \textcolor{black}{07+12} & \textcolor{black}{77.1}  & 75.6 &	88.2&	76.1&	66.7&	57.9&	86.2&	86.7&	90.5&	56.8&	88.3&	69.3&	89.9&	90.3&	79.4&	78.5&	45.6&	78.0&	78.2&	85.1&	75.6 \\
ZIP MAD \texttt{E} & 07+12 & 77.4  & 75.8&	87.5&	76.4&	67.0&	58.1&	86.6&	87.2&	90.1&	57.2&	89.1&	72.5&	87.3&	91.2&	80.1&	78.9&	46.0 &	78.5&	78.2&	85.4&	75.8
			 \\
			ZIP MAD \texttt{EB} & 07+12 & \textbf{79.8} & 79.8 & 85.5 & 78.5 & 72.1 & 57.7 & 88.8 & 88.1 & 93.1 & 58.9 & 93.7 & 73.3 & 91.8 & 92.4 & 84.9 & 81.2 & 47.5 & 81.3 & 81.5 & 88.3 & 75.5 \\
			\bottomrule
		\end{tabular}
	}
\end{table*}

\subsection{Average Precision for Object Detection}


\textbf{Comparion using the R-FCN detector.}
We follow the detection pipeline as that in the R-FCN detector \citep{rfcn}, use the ResNet-50 \citep{resNet} model and run the public code.
The batch size is 4 and the total iteration is 220,000.
Table \ref{tab:detection} reports the detection performance of different proposal methods using the \textit{same} detector at top 300 proposals - only the region proposals are different. 
It can be seen that our object proposal provides a better mAP (AP@0.5)
on the ILSVRC and are suitable for detecting objects of different sizes.
We reimplement the results of RFCN in the table using RPN proposals, which is to ensure that the AP performance difference descends from region proposals only.
It is worth noticing that although there a big gain in region proposals from ours compared with previous methods (ours vs. SS, around 30\% recall boost), the gain in terms of object detection is not that obvious (around 2\%). We believe the discrepancy  descends from the classification inaccuracy. 


\begin{table}
	\caption{Average precision (AP) for object detection on the ILSVRC \texttt{val2} set. To evaluate the performance of different proposals, we 
		use the \textit{same} R-FCN detector \citep{rfcn} across approaches. Note that the first two methods are listed for reference only since they have their own network structures and detection pipelines. 
	}
	\vspace{-.3cm}
	\begin{center}
		\footnotesize{
			\begin{tabular}{l c c c }
				\toprule
				Method & AP@0.50 & AP@0.75 & AP@0.5:0.95 \\
				\midrule
				SSD 
				& 43.4 & - & - \\
				ResNet 
				& 60.5 & - & -\\
				\midrule
				EdgeBox 
				& 49.94 & 35.24 & 31.47\\ 				
				Selective Search 
				& 51.98 & 35.13 & 34.22 \\
				AttractioNet 
				& 52.07 & 35.93 & 35.47 \\
				RFCN 
				& 53.11 & 36.41 & 35.63 \\
				ZIP & {54.08} & {36.72} & {35.66} \\
				\toprule
				Method & AP@small & AP@medium & AP@large \\
				\midrule
				EdgeBox 
				& 3.98 & 20.61 & 60.98 \\
				Selective Search 
				& 4.86 & 23.24 & 61.78 \\
				AttractioNet 
				& 4.90 & 26.37 & 63.11  \\
				RFCN 
				& 4.93& 26.81& 64.77\\
				ZIP &  {4.97} & {27.04} & {65.36} \\
				\bottomrule
			\end{tabular}
		}
	\end{center}
	\label{tab:detection}
\end{table}

\subsection{One Stage vs Two Stage}
We also embed the spirit of deconvolution and MAD unit in a single stage manner, where the SSD detector is invoked as the prototype. Compared with the 76.8\% mAP (SSD in Table \ref{tab:voc07_all}) baseline on PASCAL, the single stage trained detector with our design achieves better performance at 78.1\%, we believe the enhancement derive from better feature representation by deconvolution and the automatic feature selection via MAD unit. However, compared to our final two-stage detector, which has a 79.8\% mAP, the single stage counterpart is inferior. 
In the two-stage scheme,
region proposals are uniquely generated and separated from the second stage, which provides a more accurate set of candidate boxes (hypothesis) for the detection process in the second stage. Such an observation is in accordance with findings in \citep{trade_off_obj_det}. 

However, we still believe the single-stage spirit of object detection will dominate the field and will be prevalent. The one-stage framework is trained end-to-end without considering the two tasks separately and seems more neat and unified. We leave such an investigation as future work.